\documentclass[letterpaper]{article} 
\usepackage{aaai25}  
\usepackage{times}  
\usepackage{helvet}  
\usepackage{courier}  
\usepackage[hyphens]{url}  
\usepackage{graphicx} 
\usepackage{natbib}  
\usepackage{caption} 
\usepackage{algorithm}
\usepackage{algorithmic}

\usepackage{multirow}
\usepackage[textsize=tiny]{todonotes}
\usepackage{microtype}
\usepackage{graphicx}
\usepackage{subfigure}
\usepackage{booktabs} 
\usepackage{textcomp,booktabs}
\usepackage{amsthm}
\usepackage{booktabs}
\theoremstyle{plain}
\usepackage{xspace}
\usepackage{threeparttable}

\usepackage{amsmath}
\usepackage{amssymb}
\usepackage{mathtools}
\usepackage{amsthm}

\usepackage{tabularx}
\usepackage{amsmath}
\usepackage{pifont}

\newcommand{\xmark}{\ding{55}}
\newcommand{\ste}{{\tt S2E}\xspace}
\newcommand{\eraser}{{SEraser}\xspace}

\usepackage{newfloat}
\usepackage{listings}
\DeclareCaptionStyle{ruled}{labelfont=normalfont,labelsep=colon,strut=off} 
\floatstyle{ruled}
\newfloat{listing}{tb}{lst}{}
\floatname{listing}{Listing}
\title{Spurious Feature Eraser: Stabilizing Test-Time Adaptation \\ for Vision-Language Foundation Model}
\author{
    Huan Ma\textsuperscript{\rm 1}\equalcontrib\thanks{The project was conducted during the internship in AI Lab, Tencent with the mentor Bingzhe Wu.}, Yan Zhu\textsuperscript{\rm 1}\equalcontrib, Changqing Zhang\textsuperscript{\rm 1}\thanks{Corresponding to zhangchangqing@tju.edu.cn.}, Peilin Zhao\textsuperscript{\rm 2},\\ Baoyuan Wu\textsuperscript{\rm 2}, Long-Kai Huang\textsuperscript{\rm 2}, Qinghua Hu\textsuperscript{\rm 1}, Bingzhe Wu\textsuperscript{\rm 2}\footnotemark[2]
}
\affiliations{
    \textsuperscript{\rm 1}{College of Intelligence and Computing, Tianjin University, Tianjin, China}\\
\textsuperscript{\rm 2}AI Lab, Tencent, Shenzhen, China
}

\usepackage{bibentry}

\begin{document}

\maketitle
\begin{abstract}
Vision-language foundation models have exhibited remarkable success across a multitude of downstream tasks due to their scalability on extensive image-text paired data. However, these models also display significant limitations when applied to downstream tasks, such as fine-grained image classification, as a result of ``decision shortcuts'' that hinder their generalization capabilities. In this work, we find that the CLIP model possesses a rich set of features, encompassing both \textit{desired invariant causal features} and \textit{undesired decision shortcuts}. Moreover, the underperformance of CLIP on downstream tasks originates from its inability to effectively utilize pre-trained features in accordance with specific task requirements. To address this challenge, we propose a simple yet effective method, Spurious Feature Eraser (SEraser), to alleviate the decision shortcuts by erasing the spurious features. Specifically, we introduce a test-time prompt tuning paradigm that optimizes a learnable prompt, thereby compelling the model to exploit invariant features while disregarding decision shortcuts during the inference phase. The proposed method effectively alleviates excessive dependence on potentially misleading spurious information. We conduct comparative analysis of the proposed method against various approaches which validates the significant superiority\footnote{Codes: https://github.com/MaHuanAAA/SEraser}.

\end{abstract}

\section{Introduction}
\begin{figure}
    \centering
    \includegraphics[width=0.95\linewidth]{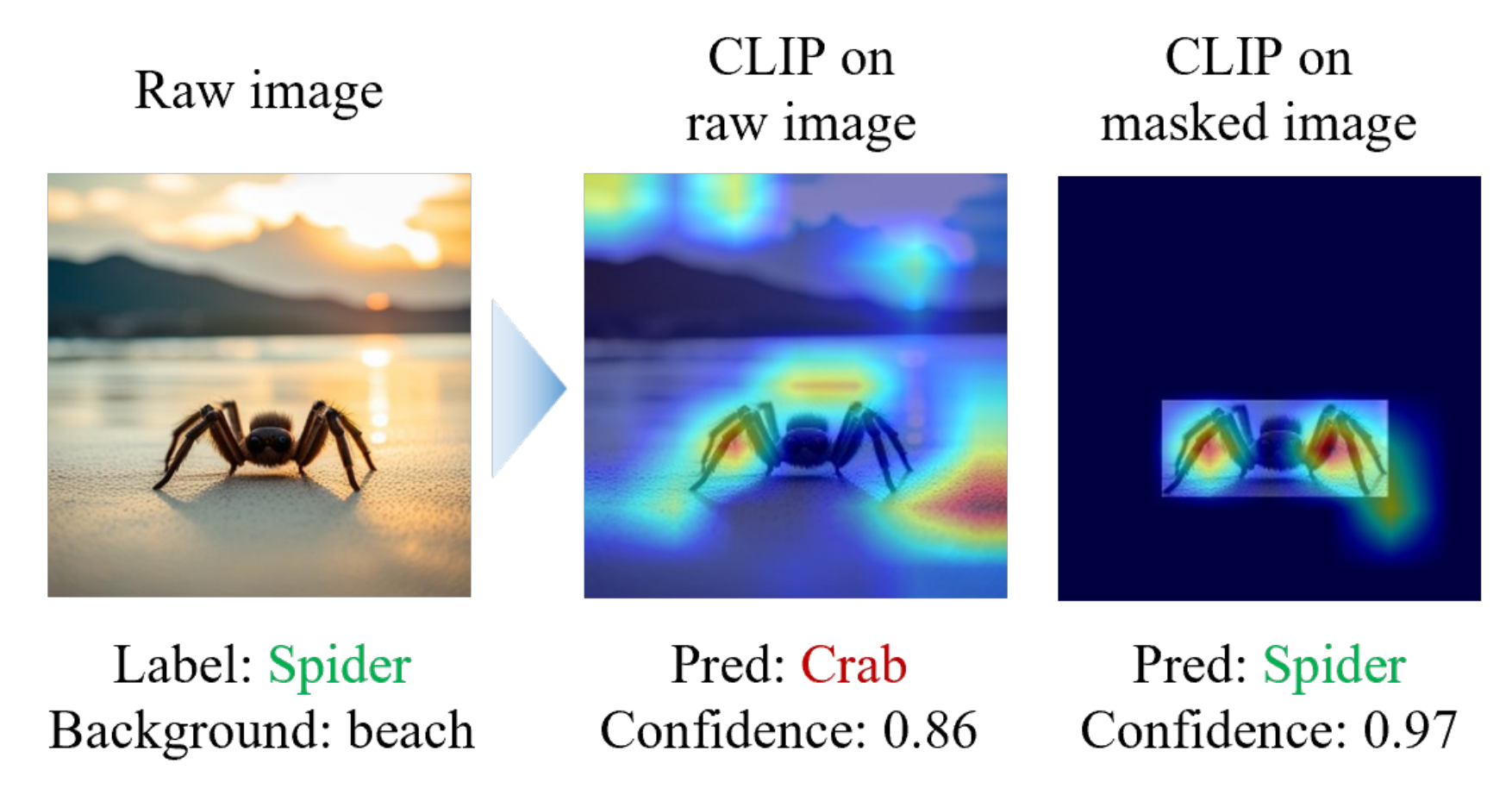}
        \caption{\textbf{Observation:} The attentive regions obtained by Grad-CAM. By eliminating beach background, CLIP changes its answer to be correct. This indicates the CLIP has learned a rich set of features (CLIP knows the object should be a spider), but the spurious features lead to decision shortcuts.
        }
    \label{fig:observation}
\end{figure}

{
\textbf{V}ision-\textbf{L}anguage \textbf{F}oundation \textbf{M}odels (VLFMs) such as CLIP~\cite{radford2021learning}, have achieved remarkable success across a diverse range of downstream tasks~\cite{chen2023ovarnet,gu2021open,xu2023open}. This success can largely be attributed to their scalability on massive image-text pair data, enabling the model to capture high-quality representation of both image and text. Such a capability further facilitates zero-shot learning on out-of-distribution (OOD) data, not limited to the original training set~\cite{kamath2021mdetr,patashnik2021styleclip,li2022grounded,ma2023fairness}.
However, despite the powerful zero-shot learning abilities of CLIP and its variants~\cite{agarwal2021evaluating,zhou2023advclip,cao2015diversity,Bai_2024_CVPR}, their application to certain downstream tasks, such as fine-grained image classification, reveals significant limitations~\cite{yang2022long,shi2023clip,ma2023calibrating}. The presence of ``decision shortcuts'' — the model's tendency to rely on simple, potentially superficial features for decision-making — severely hinders their generalization capabilities~\cite{fan2023improving,li2023whac}. These shortcuts, often emerging as a byproduct of the training process, lead to a model that, while robust in familiar scenarios, fails to effectively adapt to more fine-grained or less common tasks. 
For example, as shown in Fig.~\ref{fig:observation}, due to the existence of the shortcut of using the background for classification, the CLIP tends to predict the spider as a crab when the background is the beach.}

To improve the zero-shot generalization ability of VLFMs, several methodologies have been proposed. These methods aim to enforce the model to employ invariant features during test phases, thereby alleviating the effect of decision shortcuts. Currently, two mainstream approaches worthy of consideration are: (1) Region-aware CLIP~\cite{liang2023open,wei2023elite,sun2023alpha}: This line introduces additional region information to the CLIP, encouraging it to avoid interference from spurious features. However, a significant limitation of these methods is the necessity to finetune the CLIP's weights or even alter its original architecture~\cite{sun2023alpha}. Such alterations are costly and may also detrimentally affect the model's generalization abilities on in-distribution data due to changes in the model’s architecture and weights.
(2) Prompt Tuning~\cite{shu2022test}: This strategy involves test-time optimization of task prompts using data from downstream tasks. The optimization goal is to force the model to learn invariant representations across various augmented versions of the original visual context (e.g., rotations \& cropping), thereby mitigating the impact of visual shortcuts. Compared to the first approach, prompt tuning is advantageous as it does not require changing the original model's architecture or weights and can be applied out-of-the-box. However, these methods attempt to improve VLFMs' robustness by telling the models on which features to rely upon through various approaches, which heavily depends on the annotations of the invariant features. What's worse, forcing the model to make classification based on certain areas introduces the risk of incorporating new decision shortcuts. In this paper, we present a method which could effectively overcome the potential decision shortcuts by erasing spurious features instead of forcing the model to make classifications based on certain features, without the necessity for altering the model's architecture and weights.

Our motivation stems from the following observation: \textit{{VLFMs have already learned a rich set of features, which include both the desired invariant causal features and the unwanted decision shortcuts}}, and the reason why VLFMs fail on certain tasks is the inference process predominantly activates these decision shortcuts when applied to specific tasks. Specifically, we observe that a simple intervention by removing background information in the visual context can shift CLIP’s focus towards invariant features (the object), thereby significantly improving model performance (As shown in Fig.~\ref{fig:observation}, CLIP's zero-shot classification performance on foreground improved by $48\%$ in terms of accuracy for the worst group). In other words, this implies that if we can teach VLFMs how to overcome the impact of spurious features, their performance could be significantly enhanced.

\begin{figure*}[!t]
  \centering
      \includegraphics[width=0.94\linewidth]{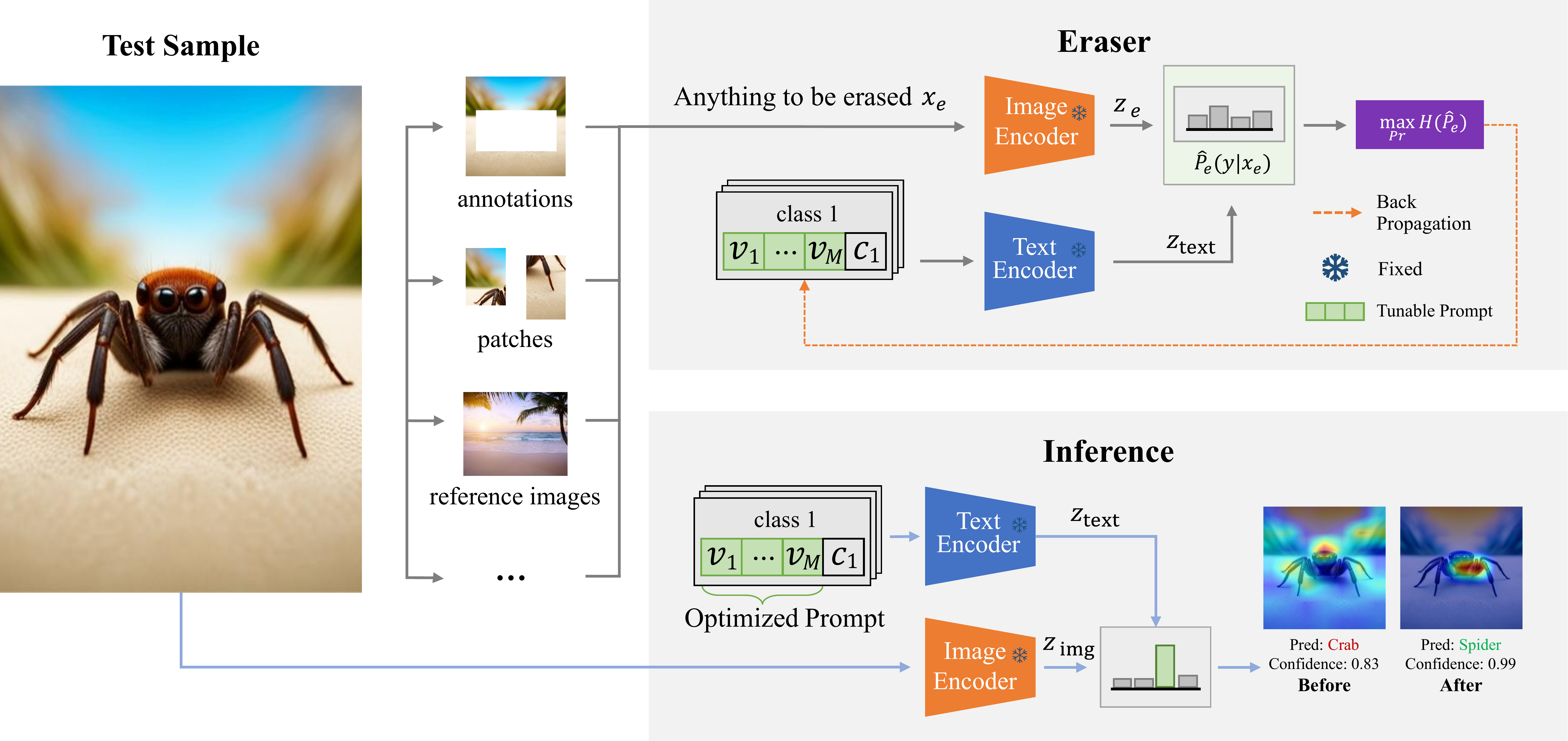}
    \caption{Framework of Spurious Feature Eraser (SEraser). Given a test sample, we first generate auxiliary images according to different strategies for test-time prompt tuning (termed as Eraser), then conduct zero-shot classification on the test sample with the optimized prompt from Eraser.}
    \label{fig:cover}
\end{figure*}

Based on this insight, we propose a test-time prompt tuning paradigm termed Spurious Feature Eraser (SEraser), encouraging VLFMs to leverage causal invariant features by disregarding any potential spurious features during inference. Unlike prior methods that directly inform VLFMs which part of the image contains invariant features, our method is more practical and flexible since it only needs to identify potential spurious features. Specifically, as shown in Fig.~\ref{fig:cover}, we sample auxiliary images first then regularize VLFMs to predict a uniform distribution on these auxiliary images through test-time prompt tuning, thereby enforcing VLFMs to learn how to ``erase'' spurious features. With the help of optimized prompt, the model can recognize spurious features, thus overcoming potential decision shortcuts inherently. Extensive experiments confirm that our approach significantly enhances the model stability against decision shortcuts compared to existing state-of-the-art methods. The core contributions of this paper are summarized as follows:
\begin{itemize}
\item For improving vision-language foundation models' performance, we propose a novel test-time prompt tuning method to overcome potential decision shortcuts by flexibly ``erasing'' the spurious features, which inherently enables vision-language foundation models to recognize and disregard spurious features. 
\item We develop a new evaluation paradigm for vision-language foundation models, and utilize it as a supplementary benchmark to assess VLFMs' reliability. This evaluation framework effectively reflects the ability of overcoming decision shortcuts for VLFMs. And it can avoid data leakage, which may lead to unfair evaluation. 
\item Our method significantly improves the zero-shot classification performance of VLFMs, especially when they exhibit clear decision shortcuts on that task. As the empirical result in the benchmark dataset Waterbirds~\cite{koh2021wilds}), our method can even bring over $25\%$ improvement on the worst group.
\end{itemize}

\section{Method}

{In this section, we first introduce the problem setting and corresponding notations. Building upon the observation that ``decision shortcuts'' hinder the generalization capabilities of VLFMs, we introduce the Spurious Feature Eraser (SEraser) to improve the reliability of current VLFMs. Furthermore, we show the reasons why previous evaluation paradigms are not suitable for VLFMs, and consequently propose a new evaluation paradigm.}

\subsection{Notations}
In the context of a vision-language foundation model $\mathcal{M}$, such as CLIP, we explore its application in downstream tasks, particularly in zero-shot scenarios. For a given test image sample $x$ with a potential label set $\mathcal{Y}=\{1,2,\cdots,K\}$, zero-shot classification requires using $\mathcal{M}$ to categorize $x$ into its appropriate class, denoted as $\hat{y}\in \mathcal{Y}$.
Specifically, each category-specific text input is prefixed with a hand-crafted prompt, for example, ``{\tt a photo of a}'', leading to class descriptions like ``{\tt a photo of a waterbird}''. These descriptions are then transformed into their respective embeddings $z_\text{text}^{k},~k=1,\cdots,K$, using the text encoder. Simultaneously, the image $x$ is processed through the image encoder, resulting in its image embedding $z_\text{img}$.
Subsequently, the normalized prediction distribution $\hat{P}(y|x)={p(\hat{y}=k|x)}_i^K$ is computed using the softmax function, which is based on the cosine similarity between the image and text: $ p(\hat{y}=k|x) = \frac{\exp(\cos(z_\text{text}^{k},z_\text{img})/\tau)}{\sum_{j}^{K}\exp(\cos(z_\text{text}^{j},z_\text{img})/\tau)}$, 
where $\cos(\cdot,\cdot)$ and $\tau$ indicate the cosine similarity between two embeddings and temperature, respectively.

\subsection{Observation: CLIP Contains Rich Features}

Based on the testing protocol  outlined above, we craft a series of tasks to assess the reliability of the CLIP when confronted with decision shortcuts. Through comprehensive evaluation, we arrive at two core conclusions: (1) The CLIP, during its pre-training phase, has acquired a rich set of features. This set encompasses stable causal attributes that genuinely benefit downstream tasks, as well as spurious decision shortcuts that lead to erroneous associations.
(2) In the unconstrained zero-shot inference stage, CLIP is highly susceptible to contextual influences, often resorting to the use of decision shortcuts for classification, so we can teach VLFMs how to overcome the impact of spurious features to improve their performance.

To illustrate this, we employ the SpiderCrab dataset as an example.
In the classification task of differentiating crabs from spiders, our initial observations reveal markedly poor reliability of the CLIP in the most challenging test subgroup, which involves spiders against a beach background. The accuracy in this subgroup is strikingly low, hovering around only $42\%$. Utilizing visualization tools like Grad-CAM (Fig.~\ref{fig:observation}), we further find that the model's focus was predominantly on background-related decision shortcuts, aligning with our second conclusion.
However, a significant improvement is noted when we employ SAM model~\cite{wang2023sam} to extract the foreground context containing the core object as the model's input. This alteration leads to a substantial increase in accuracy, soaring to $90\%$, thereby supporting our first conclusion.

\subsection{Spurious Feature Eraser}
The experimental observations prompt us to contemplate how to compel foundational models to focus on and utilize causal features. In the following sections, we introduce a novel test-time adaptation paradigm that achieves this objective without modifying the original model's structure or weights. We begin by presenting the basic methodology of visual-language prompt tuning. Subsequently, we will introduce our proposed approach and the details of how to construct auxiliary images indicating spurious features.

Prompt tuning is an efficient and lightweight fine-tuning strategy specifically designed for foundation models \cite{liu2022p}. For VLFMs, prompt tuning operates by fine-tuning the prompt $Pr$ of the sentences~\cite{zhu2023prompt}. Specifically, the prompt is initialized by a hand-draft sentence, such as ``{\tt a photo of a [class$_{k}$]}'', which will be tokenized as a vector $t_k=\{v_1, \cdots, v_{M}, c_k\}$, then we fix the class token $c_k$ and fine-tune a set of $M$ continuous context vectors $Pr:=\{v_1, \cdots, v_{M}\}$: 
\begin{equation}
    Pr^* = \arg \min_{Pr}\mathcal{L}(\mathcal{M}, Pr, x),
    \label{eq:prompttuning}
\end{equation}
where $\mathcal{L}$ is the optimization goal.

Prior work \cite{shu2022test} has further leveraged the concept of minimizing entropy optimization to design an appropriate objective function, enhancing the model's reliability through test-time adaptation of prompts. Specifically, for a given test sample, the prompt is optimized by minimizing the entropy of averaged prediction distribution over augmented views and filters out the augmented views with low confidence to discard spurious features. However, these methods focus on instructing the models on which features to depend upon and heavily rely on the annotations of the invariant features. Consequently, they tend to fail when the annotations are not exactly precise.

 \begin{figure*}[th]
    \centering
    \includegraphics[width=0.98\linewidth]{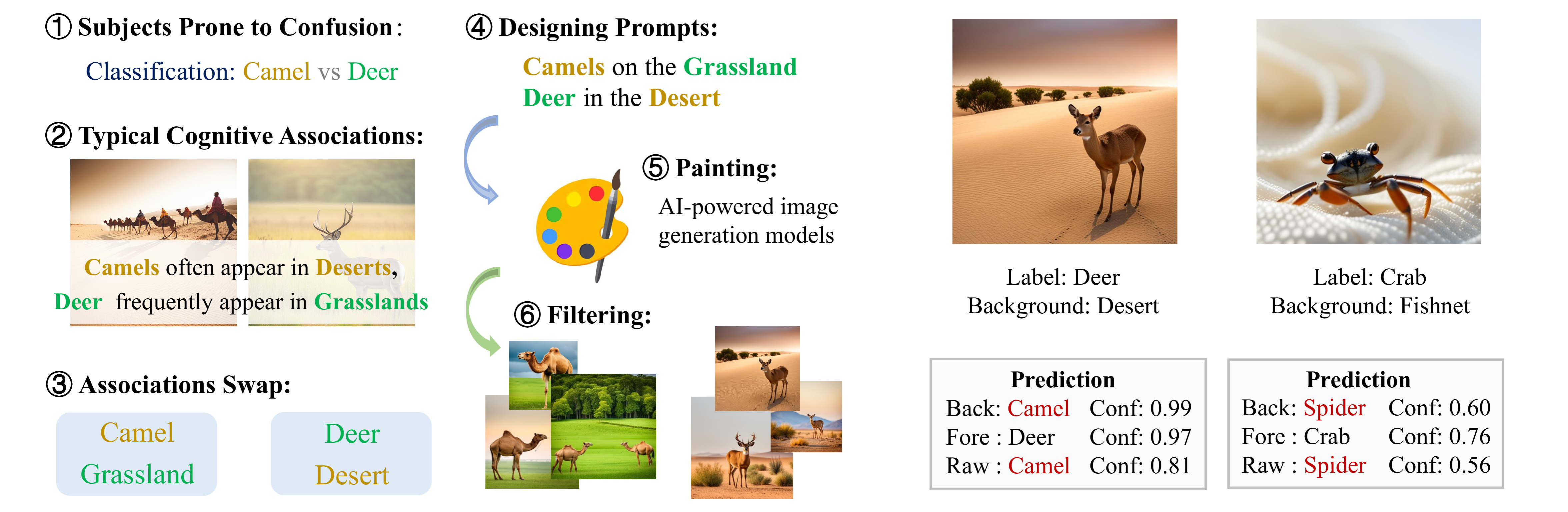}

        \caption{Pipeline of \ste and the predictions over classes of CLIP on different images. We can find that CLIP can correctly classify the objects on foreground, but its predictions on the images with background tend to be wrong when there are serious decision shortcuts on the background.}
    \label{fig:painting}
\end{figure*}

The presence of spurious features in data often involves potential decision shortcuts, which can inadvertently influence models' performance. To mitigate this, our approach aims to minimize the impact of such spurious features. We achieve this by optimizing the test prompts through maximizing the entropy of the predictive distribution for these potential spurious features, effectively steering it towards a more uniform distribution. Specifically, suppose we want to erase the spurious features in an auxiliary image $x_\text{e}$, we minimize the Kullback-Leibler divergence between the prediction distribution on the auxiliary image $\hat{P}_\text{e}$ and the uniform distribution $q$:
\begin{equation}
    \mathcal{L}_\text{e} =\text{KL}(\hat{P}_\text{e}(y|x_\text{e})\|q),
\label{eq:loss_ir}
\end{equation}
where $\text{KL}(\cdot)$ indicates Kullback-Leibler divergence. The process is shown in Fig.~\ref{fig:cover}, where the loss in Eq.~\ref{eq:loss_ir} is used as the optimization objective to achieve the removal of decision shortcuts in VLFMs on the input image. Next, we will introduce in detail how to construct auxiliary images.

\subsection{Construction of Auxiliary Images}

The method proposed in this paper exhibits a high degree of flexibility, allowing for various choices in constructing auxiliary images. In this paper, we primarily introduce three simple yet effective strategies as examples (illustrated in Fig.~\ref{fig:cover}). 
$\bullet$~\textbf{\underline{Annotations}}: Assuming that we can obtain annotations of spurious features, such as expert annotations in medical imaging and background area annotations in target recognition, we can implement them as the features to be erased by SEraser, thereby achieving precise erasure of spurious features. In this paper, we simulate an expert's annotation by employing Segment Anything~\cite{kirillov2023segment} to annotate the background.
$\bullet$~\textbf{\underline{Patches}}: In some tasks, obtaining the annotations of spurious features is difficult. Fortunately, we can still utilize alternative choices as auxiliary images. For instance, we can sample small patches from input images through multiple strategies as proxies, based on the assumption that the model should not make hasty judgments relying on partial information.
$\bullet$~\textbf{\underline{Reference images}}: Drawing inspiration from work utilizing reference data to enhance model robustness, we sample similar images from reference datasets as auxiliary images. For example, in a bird classification task, we can select images from an aircraft classification dataset as auxiliary images for SEraser.

\textbf{Flexibility of Auxiliary Image Construction.} If the auxiliary images are strictly prohibited from containing any invariant features, it would severely harm the flexibility of the proposed method. However, the approach remains effective even when these images do contain such invariant features, as it still improves the robustness of VLFMs. The method operates as a soft constraint, suggesting that decisions should not be solely based on certain features, as opposed to a rigid constraint that forces decisions to be based on specific features. \textit{The flexibility enables the method to be more adaptable, and even when the auxiliary images contain partial invariant features, it is still effective since it prevents the model from making decisions based on partial features}. The similar strategy is also used in OOD detection~\cite{ID-Like}, which improves OOD detection performance using ID-like features. Therefore, the method can still ensure trustworthy decision even when auxiliary images contain invariant features. Moreover, we can design a more reasonable approximation of spurious features in auxiliary images using multiple strategies, thereby jointly improving the trustworthiness of decision. Ablation study demonstrates that even if partial invariant features in the image are erased, the proposed method can still enhance the robustness of VLFMs, resulting in a better performance.

In experiments, we find that current testing protocols and benchmarks, such as WILDS~ \cite{koh2021wilds}, are no longer suitable for VLFMs. These benchmarks are no longer as effective in evaluating VLFMs as they are for models such CNN models, and various issues arise. In the following section, we will discuss in detail why these benchmarks are no longer suitable and propose a new robustness evaluation paradigm for VLFMs to address these shortcomings.

\begin{table*}[ht]
\centering

{

\begin{tabularx}{0.99\linewidth}{
    m{.08\linewidth}
    m{.56\linewidth} 
    m{.11\linewidth} 
    m{.12\linewidth}}

\toprule
\textbf{Method} 
& \textbf{Description}  
& \textbf{Augment} 
& \textbf{\quad Tuning} 
\\\midrule

{Vanilla}    
& basic comparative method, the original CLIP model is deployed for zero-shot classification directly.
& None  
& \quad\quad\xmark    
\\\midrule

{MASK}    
& using SAM~\cite{kirillov2023segment} to annotate the foreground and mask the background, then the masked image is classified in a zero-shot classification manner.
& background annotation   
& \quad\quad\xmark  
\\\midrule

{TPT}    
& a \textbf{\emph{{prompt tuning}}} method optimizes the prompt by minimizing the entropy of averaged prediction distribution over augmented views with confidence selection ~\cite{shu2022test}. 
& views augmentation
& \,\, prompts  
\\\midrule

{RoSHOT} 
& a method (ROBOSHOT) which uses LLMs to obtain useful insights from task descriptions, then these insights are embedded and used to remove harmful and boost useful components in embeddings ~\cite{adila2023zero}.
& positive \& negative insights
& \quad logits
\\\midrule

{$\alpha$-CLIP} 
& a \textbf{\emph{{region-aware}}} version of CLIP with an auxiliary alpha channel to suggest attentive regions and fine-tuned with constructed millions of RGBA region-text pairs ~\cite{sun2023alpha}. 
& background annotation    
& architecture and weights  
\\ \midrule
{SEraser (Ours)} 
& a test-time prompt tuning paradigm that optimizes a learnable prompt, helping the model to disregard decision shortcuts during the inference phase.
& features to be erased \, (optional)
& \,\, prompts
\\
\bottomrule

\end{tabularx}}
\caption{Descriptions of different methods compared in experiments.}
\label{tab:existing_methods}
\vspace{-3mm}
\end{table*}

\section{{\ste}: Evaluation Protocol} 
In this section, we present a novel evaluation paradigm for assessing VLFMs. We find that current testing protocols and benchmarks, such as WILDS~ \cite{koh2021wilds,han2022trusted}, are no longer suitable for VLFMs, because they are built through large-scale self-supervised learning, and it will lead to \textit{(1) data distribution shift in VLFMs paradigm no longer same as benchmarks} and \textit{(2) testing data leakage}.

(1) In this context conventional robustness testing, most decision shortcuts are caused by carefully designed training datasets (i.e., distribution shifts in training data). A commonly employed benchmark is the WILDS dataset collection proposed by Stanford University \cite{koh2021wilds}. However, these data-centric robustness generalization testing protocols and benchmarks have become unsuitable for VLFMs developed through large-scale self-supervised learning. Specifically, decision shortcuts in traditional robustness testing benchmarks stem from carefully designed on limited-scale training data (e.g., decision shortcuts between image style and label designed in the PACS benchmark~\cite{xu2019self}), whereas VLFMs are built using large-scale open-source data, rendering \textit{{these designed decision shortcuts often absent in VLFMs}}.

(2) During the evaluation process of foundation models using benchmarks, a significant issue is \textit{{test data leakage}}, and the same problem has been highlighted in LLM evaluation~\cite{zhou2023don}. These foundation models are trained on large-scale web data, and many samples in testset themselves or highly similar samples are present in the training data. For example, we find that alpha-CLIP performs exceptionally well on benchmarks such as Tiny-ImageNet. However, from the training details of alpha-CLIP, we find the released checkpoints are a retrained version of CLIP using ImageNet, implying its superior performance on the ImageNet relevant benchmarks. Similar issues of test data leakage in LLM evaluation are increasingly being acknowledged, and new simulated data are being generated to ensure the validity of model evaluation.

To bridge the existing gap in this field, this paper proposes a novel paradigm for evaluating the reliability of VLFMs, termed ``shortcut-to-evaluate'' ({\ste}). The overall pipeline is illustrated in the left portion of Fig.~\ref{fig:painting}. Specifically, the process can be described as the following steps: (1) We identify a set of subjects that typically challenge the model's classification accuracy, such as differentiating between camels and deer. (2) We then examine their common cognitive associations - camels are typically associated with desert environments, while deer are often linked to grassland habitats. (3) Then, we interchange these environmental associations, positioning camels in grassland settings and deer in desert landscapes. This step is pivotal in creating decision-making shortcuts for the model to follow. (4) We refine these unusual combinations into specific instructions designed for advanced AI-powered image generation tools, such as DALL-E and Midjourney. (5) Using these customized instructions, we generate a series of test images, each crafted to evaluate the model's ability to handle these unconventional scenarios. (6) The final step involves a rigorous filtering process, discarding any images that do not meet our predefined criteria, such as those lacking the presence of the intended animals. In this paper, we generate two scenarios according to \ste.

\section{Experiments}

\subsection{Setup}
In the experiments, we evaluate different methods on different scenarios, including the real-word image data Tiny-ImageNet~\cite{le2015tiny}, CUB-200~\cite{Wah2011TheCB} and the benchmark simulated data {Waterbirds}~\cite{koh2021wilds}, and the datasets created by \ste CamelDeer and SpiderCrab. To intuitively demonstrate the decision shortcuts in VLFMs, we primarily focus on the samples of challenging classes in the dataset (for details please refer to Appendix). Except for Alpha-CLIP, which has its own separately designed visual encoder, all methods use the CLIP with a pre-trained ViT-B-32 released by OpenAI~\cite{radford2021learning} as the visual encoder, which is a representative architecture of vision encoder~\cite{conde2021clip,wu2023tinyclip,chen2023disco}.

\begin{table*}[ht]
  \centering

  \begin{threeparttable}[b]{
    \begin{tabular}{c|cccc|ccc}
    \toprule
    {Dataset} & {Vanilla}& TPT & RoSHOT & {$\alpha$-CLIP}\tnote{$\ast$} & Patches  &   Images   & Both\\
    \midrule 
   {Tiny-ImageNet}
    &$23.2$ &$29.6$ &$\pmb{49.2}$ &${{76.0}}$ &$42.4$ &$41.2$ &${42.8}_{{\scriptstyle(\blacktriangle19.6)}}$\\
    \midrule 
    {CUB-200}
    &$12.1$ &$8.7$ &$25.5$ &${{44.3}}$ &$26.2$ &$24.2$ &$\pmb{28.9}_{{\scriptstyle(\blacktriangle16.8)}}$\\
    \midrule 
    {ImageNetA}
    &$42.1$ &$\pmb{49.7}$ &$38.9$ &${{51.5}}$ &$47.4$ &$45.4$ &$\pmb{49.7}_{{\scriptstyle(\blacktriangle7.6)}}$\\
    \midrule 
    \midrule 
    {Average}
    &$25.8$ &$29.3$ &$37.9$ &${{57.3}}$ &$38.7$ &$36.9$ & $\pmb{40.5}_{{\scriptstyle(\blacktriangle14.7)}}$\\

    \bottomrule
    \end{tabular}}%
   \begin{tablenotes}
    \small
     \item[$\ast$] \textit{Clarification: {$\alpha$-CLIP was trained on multiple open-source datasets including ImageNet}, which renders the above evaluation unsuitable for this model.} 
   \end{tablenotes}
      \end{threeparttable}
    \caption{Zero-shot classification performance for different choices of how to construct auxiliary images on real-world scenarios, where ``Average'' indicates the averaged performance on all datasets and the {$\blacktriangle$Green} mark represents the improvement relative to Vanilla. }\label{tab:realworld}
\end{table*}

\begin{table*}[ht]
  \centering

  \begin{threeparttable}[b]
{
    \begin{tabular}{cc|cccccccc}
    \toprule
    \multicolumn{2}{c}{Dataset} & {Vanilla}&MASK  &{TPT} 
    & RoSHOT & $\alpha$-CLIP & {Ours}\\
    \midrule 

    \multirow{2}{*}{Waterbirds}
    &$\text{AVG.}$ &$67.67$ &$71.97$ &$66.88$ &$68.86$ &$67.59$ &$\pmb{78.24}_{{\scriptstyle(\blacktriangle10.57)}}$ \\
    &$\text{W.G.}$  &$40.04$ &$51.53$ &$34.38$ &$52.28$ &$43.15$ &$\pmb{65.25}_{{\scriptstyle(\blacktriangle25.21)}}$\\
    \midrule 

     \multirow{2}{*}{CamelDeer\tnote{$\ast$}}
    &$\text{AVG.}$ &$83.20$ &$93.60$ &$77.67$ &$80.40$ &$92.00$ &$\pmb{95.67}_{{\scriptstyle(\blacktriangle12.47)}}$\\
    &$\text{W.G.}$ &$66.40$ &$87.20$ &$55.33$ &$60.80$ &$84.40$ &$\pmb{91.60}_{{\scriptstyle(\blacktriangle25.2)}}$\\
    \midrule 
    \multirow{2}{*}{SpiderCrab\tnote{$\ast$}}
    &$\text{AVG.}$ &$66.00$ &$91.40$ &$83.53$ &$73.00$ &$86.20$ &$\pmb{95.33}_{{\scriptstyle(\blacktriangle29.33)}}$\\
    &$\text{W.G.}$ &$42.00$ &$90.40$ &$72.53$ &$50.40$ &$86.00$ &$\pmb{94.67}_{{\scriptstyle(\blacktriangle52.67)}}$\\
    
    \midrule 
    \multirow{2}{*}{Average}
    &$\text{AVG.}$ &$72.29$ &$85.67$ &$76.03$ &$74.09$ &$81.93$ &$\pmb{89.75}_{{\scriptstyle(\blacktriangle17.46)}}$\\
    &$\text{W.G.}$ &$49.48$ &$76.38$ &$54.08$ &$54.49$ &$71.18$ &$\pmb{83.72}_{{\scriptstyle(\blacktriangle34.24)}}$\\
    
    \bottomrule
    \end{tabular}}%
    \begin{tablenotes}
    \small
     \item[$\ast$] We submit a subset of these datasets in supplementary materials limited by the maximum file size. 
   \end{tablenotes}
      \end{threeparttable}
       \caption{Zero-shot classification performance on simulated scenarios, where ``AVG.'' and ``W.G.'' indicate the average performance on the entire test set and the worst-performing group (i.e., the group with the lowest accuracy), respectively (for error bars please refer to Appendix).}\label{tab:main_tab}
\end{table*}

\subsection{Main Results}

\subsubsection{Results on Real-world Scenarios}

We evaluate the performance of the proposed method on real-world datasets, and the results are presented in Table~\ref{tab:realworld}. ``Patches'' denotes the strategy of cropping patches of the test image to serve as auxiliary images, while ``Images'' refers to the approach of employing reference images as auxiliary images. ``Both'' indicates jointly employing both strategies in deploying SEraser. Specifically, \textbf{\underline{Patches}} strategy divides the input image into 8x8 patches and utilizes the four corner patches as auxiliary images for SEraser implementation to improve the diversity of auxiliary images (adjacent patches often exhibit lower information complementarity~\cite{wang2023sammed}). On the other hand, the ``Images'' involves selecting images from benchmarks that do not belong to any category in the classification task as OOD datasets and searching for the most similar images based on cosine similarity to serve as \textbf{\underline{Reference images}}. ``Both'' represents the deployment of SEraser using auxiliary images obtained from the aforementioned strategies (for further details, please refer to Appendix).
The results demonstrate that different strategies improve zero-shot classification performance, with the combined application of both strategies yielding higher performance than using either strategy independently. It is worth noting that Alpha-CLIP ($\alpha$-CLIP) achieves outstanding performance on the open-source benchmark Tiny-ImageNet, which can be attributed to the checkpoint being retrained on ImageNet. 

\subsubsection{Results on Simulated Scenarios}

We conduct experiments using the Waterbirds benchmark as well as simulated tasks generated through the \ste. We adapt SAM~\cite{kirillov2023segment} for image segmentation, which identifies the background as auxiliary images by creating bounding boxes (i.e., simulated an expert's \textbf{\underline{Annotations}}). If an image contains multiple foreground boxes, we combine them as a single foreground. Furthermore, since only one test image is input in a zero-shot manner prompt during test-time adaptation, we apply regularization on the model to minimize the prediction distribution entropy on features that are not to be erased and employ  RandAugment~\cite{cubuk2019randaugment} on the images to prevent trivial solutions. Both Alpha-CLIP (region-aware) and our proposed method substantially enhance the performance of the CLIP, particularly in the worst group. Alpha-CLIP benefits from extensive data-driven region-aware fine-tuning and the annotation information of spurious features during the testing phase, resulting in a markedly improved performance compared to the original CLIP. On the other hand, our method achieves the best performance through test-time adaptation, without finetuning the model's weights. The performance of TPT is suboptimal due to its test-time adaptation objective, which minimizes the average entropy by randomly cropping the input image. TPT assumes that the views with higher confidence represent invariant features, while those with lower confidence are spurious features. However, this assumption is occasionally unsuitable. As illustrated in Fig.~\ref{fig:painting}, higher confidence in the desert background than in the foreground  leads the TPT models to make decisions based on the background. RoSHOT exhibits improvement in some tasks but a decline in others.  We can observe that our method has achieved significant improvements across various datasets, particularly in the aspect of worst group performance.


\subsection{Ablation Study}

\begin{table}[ht]
  \centering

{
  \resizebox{\linewidth}{!}{
    \begin{tabular}{cc|ccc}
    \toprule
    {Models}& & {Vanilla}&MASK & {Ours}\\
    \midrule 

\multirow{2}{*}{CLIP-L14}&$\text{AVG.}$ 
&$83.71$ &$85.48$  &$\pmb{87.78}_{{\scriptstyle(\blacktriangle4.07)}}$\\
&$\text{W.G.}$  &$32.87$ &$40.81$  &$\pmb{58.88}_{{\scriptstyle(\blacktriangle26.01)}}$\\
    \midrule 

\multirow{2}{*}{BLIP-2}&$\text{AVG.}$ 
&$\pmb{57.65}$ &$54.38$  &${55.56}_{{\scriptstyle(\blacktriangledown2.09)}}$\\
&$\text{W.G.}$  &$28.19$ &$35.05$  &$\pmb{34.74}_{{\scriptstyle(\blacktriangle6.55)}}$\\
\midrule 
\midrule 
\multirow{2}{*}{Average}&$\text{AVG.}$ 
&$70.68$ &$69.93$  &$\pmb{71.67}_{{\scriptstyle(\blacktriangle0.99)}}$\\
&$\text{W.G.}$  &$30.53$ &$37.93$  &$\pmb{46.81}_{{\scriptstyle(\blacktriangle16.28)}}$\\

    \bottomrule
    \end{tabular}}}%
  \caption{Zero-shot classification performance of different VLFMs on Waterbirds. }\label{tab:othermodels}

\end{table}

\textbf{Effectiveness Across Different Model Architectures.} The results presented before demonstrate that the proposed paradigm significantly enhances the zero-shot classification capability of CLIP ViT-B-32. To evaluate the method's flexibility across other models, we evaluate its performance on other widely-used architectures, including CLIP ViT-L-14 and BLIP-2. We conduct these evaluations using the Waterbirds benchmark, a widely employed benchmark dataset. As shown in Table~\ref{tab:othermodels}, our method achieves promising performance on both models. Specifically, the CLIP with ViT-L14 image encoder is improved significantly after deploying \eraser, and BLIP-2 shows a significant improvement on the worst group, validating the effectiveness of \eraser across diverse architectures (for results on other datasets, please refer to Appendix).

\begin{figure}[t]
\centering
  \includegraphics[width=0.70\linewidth]{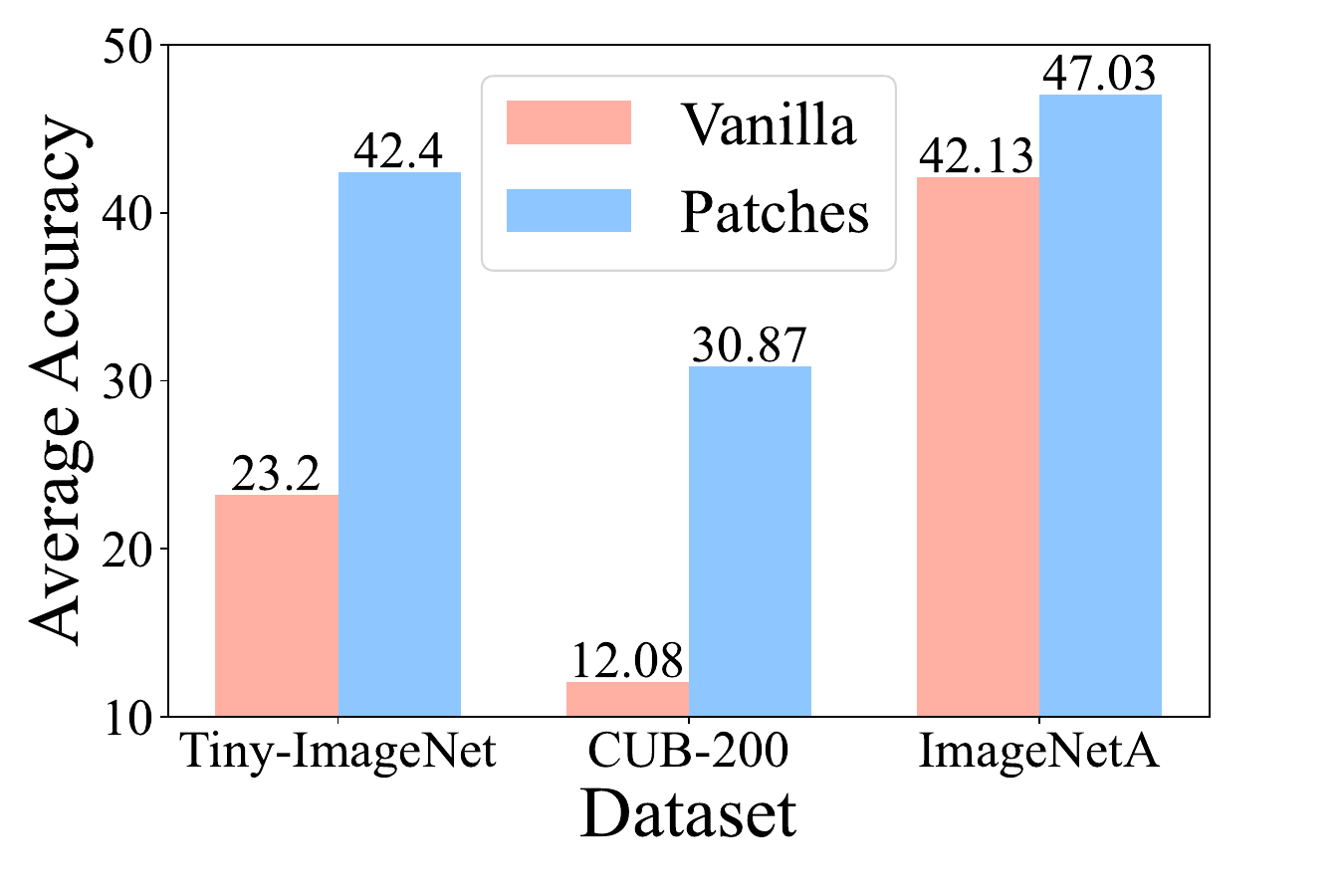}
 \caption{Performance under different patch sampling strategies.}

 \label{fig:random}
\end{figure}
\textbf{Flexibility in Constructing Auxiliary Images.} The flexibility of the proposed method in the presence of invariant information in auxiliary images is examined through the following ablation study. We modify the sampling strategy of \textbf{\underline{Patches}} to investigate this aspect. Specifically, we replace the corner patches with a random patch sampling strategy. This approach implies that the patch sampling is entirely random, without any prior knowledge, and thus, it can sample patches that contain invariant features and are adjacent. As illustrated in Fig.~\ref{fig:random}, even when SEraser is implemented using these randomly sampled patches, the proposed method still enhances the performance of VLFMs. This improvement can be attributed to the soft constraint proposed in SEraser, which ensures that ``VLFMs cannot make hasty decisions based on partial information'', accordingly, assisting VLFMs in alleviating decision shortcuts on then patches even containing partial invariant features.

\section{Related Work}

$\bullet$~\textbf{Region-aware CLIP}: One straightforward strategy is to {mask the background} \cite{liang2023open}, forcing the model to focus on the foreground. Some other approaches prompt the CLIP model by {circling the foreground} of the input image, guiding CLIP to focus on the area of interest. For example, Red-Circle~\cite{shtedritski2023does} and FGVP~\cite{yang2023fine} utilize a circle or mask contour to indicate where CLIP should concentrate its attention. 
{Alpha-CLIP}~\cite{sun2023alpha} enhances CLIP by incorporating regions of interest through an additional alpha channel input. This supplementary channel is derived using SAM, which ranges from [0, 1], where 1 represents the foreground and 0 represents the background.
$\bullet$~\textbf{Prompt Tuning}: Test-time Prompt Tuning~\cite{shu2022test} based on {view-augmentation} optimizes the prompt to encourage consistent predictions across augmented views by minimizing the marginal entropy and filtering out noisy augmentations with low confidence. However, it is important to note that this assumption may not always hold true. For instance, in an image containing a landbird and water background, spurious features (water background) can actually lead to high confidence rather than predicting a uniform distribution.
ROBOSHOT~\cite{adila2023zero} uses zero-shot to obtain useful insights from task descriptions. Instead of to find a optimized prompt, the authors utilize the prompts generated by LMs to push the model focus on invariant features. The insights generated from LMs are embedded and used to remove harmful and boost useful components in embeddings.
PerceptionCLIP ~\cite{an2023more} finds that providing CLIP with contextual attributes improves zero-shot classification and reduces reliance on spurious features, and \cite{chuang2023debiasing} measure spurious correlations through biased prompts and then debias the model using orthogonal projection.

\vspace{0.3cm}

\section{Conclusion}
In this paper, we propose a method to address the challenges faced by vision-language foundation models in decision shortcuts. Starting from the observation that CLIP contains both desired invariant causal features and undesired decision shortcuts, we find that the latter is responsible for the underperformance in specific tasks. Then, a test-time prompt tuning paradigm is introduced to overcome the possible decision shortcuts, which optimizes a learnable prompt to encourage the model to focus on genuine causal invariant features while restraining decision shortcuts in inference. A comparative analysis is conducted to validate the effectiveness of the proposed method against existing approaches.

\section{Acknowledgments}

This work is supported by the National Natural Science Foundation of China Grant No.62376193, Grant No.61925602, and Grant No.U23B2049.

\bibliography{aaai25}

\end{document}